\RequirePackage{fix-cm}
\documentclass[smallextended]{svjour3}       
\smartqed  
\usepackage{graphicx}
\usepackage{placeins}
 \journalname{Journal of Automated Reasoning}

\title{Finding  Proofs in Tarskian Geometry}

\author{Michael Beeson        \and
        Larry Wos 
}

\institute{Michael Beeson \at
              San Jos\'e State University \\
              \email{profbeeson@gmail.com}           
           \and
           Larry Wos \at
              Argonne National Laboratory \\
             \email{wos@mcs.anl.gov}
}

\date{Received: date / Accepted: date}

\usepackage{pstricks}
\usepackage{epsf}
\usepackage{amsfonts,amssymb}
\usepackage{makeidx}  
\usepackage{upquote}  

\usepackage{graphicx}
 
\psset{unit=3cm}
\def\Otter{{\textsc OTTER\ }}
\def\Ott{{\textsc OTTER}}  

\def\axioms{\begin{tabular*} {0.7\textwidth}{@{\extracolsep{\fill}}ll}}
\def\endaxioms{\end{tabular*}\smallskip}
\def\implies{\ \rightarrow\ }

\def\F{{\mathbb F}}

\def\B{{\bf B}}
\def\T{{\bf T}}

\def\InnerOuterPaschFigure{%
\pspicture(3.6,1.7)
\psdot(0,0.6)
\put(0,0.47){$a$}
\pscircle(1,0.6){0.03}
\put(1,0.47){$x$}
\qline(0,0.6)(0.97,0.6)
\psdot(1.3,0.6)
\put(1.35,0.55){$q$}
\qline(1.03,0.6)(1.3,0.6)
\psdot(1.1,1.4)
\put(1.1,1.45){$c$}
\qline(0,0.6)(1.1,1.4)  
\qline(1.1,1.4)(1.375,0.3) 
\psdot(1.375,0.3)
\put(1.43,0.25){$b$}
\psdot(0.52,0.98)
\put(0.43,1.03){$p$}
\qline(1.375,0.3)(1.015,0.584)  
\qline(0.52,0.98)(0.979,0.62)  
\psdot(2,0.3)
\put(1.95,0.19){$a$}
\psdot(3.5,0.3)
\put(3.45,0.19){$q$}
\qline(2,0.3)(2.47,0.3)
\qline(2.53,0.3)(3.5,0.3)
\pscircle(2.5,0.3){0.03}
\put(2.45,0.19){$x$}
\psdot(2.7,1.5)
\put(2.65,1.57){$b$}
\qline(2.7,1.5)(3.5,0.3)  
\psdot(3.0,1.05)
\put(3.07,1.02) {$c$}
\qline(2.7,1.5)(2.5,0.33)  
\qline(2,0.3)(3.0,1.05)   
\psdot(2.566,0.722)
\put(2.6,0.6){$p$}
\endpspicture}

\def\TarskiParallelFigure{%
\pspicture(2.5,1.2)(0,-0.15)
\qline(0.03,0)(2.47,0)  
\pscircle(0,0){0.03}
\put(-0.13,-0.02){$x$}
\pscircle(2.5,0){0.03}
\put(2.55,-0.02){$y$}
\psdot(0.5,1)
\put(0.47,1.05){$a$}
\psdot(0.167,0.333)
\put(0.06,0.33){$b$}
\qline(0.5,1)(2.48,0.022)  
\qline(0.01,0.025)(0.5,1)  
\psdot(1.1,0.703)
\put(1.1,0.74){$c$}
\qline(0.167,0.333)(1.1,0.703)  
\qline(0.5,1)(1.0,0)  
\psdot(1.0,0)
\put(1.02,0.05){$t$}
\psdot(0.723,0.553)
\put(0.78,0.48){$d$}
\endpspicture
}
 
\def\TarskiFiveSegmentFigure{%
\hskip 0.5in
\psset{unit=2.25cm}
\pspicture(2.2, 0.8)
\qline(0.0,0.15)(2.0,0.15)
\qline(0.0,0.15)(1.0,0.85)
\psline[linestyle=dashed](1.0,0.85)(2.0,0.15)
\qline(1.0,0.85)(0.7,0.15)
\put(1,0.9){$d$}
\put(0.0,0){$a$}
\put(0.7,0){$b$}
\put(2.0,0){$c$}
\qline(2.5,0.15)(4.5,0.15)
\qline(2.5,0.15)(3.5,0.85)
\psline[linestyle=dashed](3.5,0.85)(4.5,0.15)
\qline(3.5,0.85)(3.2,0.15)
\put(3.5,0.9){$D$}
\put(2.5,0){$A$}
\put(3.2,0){$B$}
\put(4.5,0){$C$}
\psset{unit=3cm}
\endpspicture}

\def\CrossBarFigure{%
\pspicture (5,1.75)(0.5,0)
\psdot(2.25,0.25)
\put(2.25,0.14){$a$}
 \put(4,0.14){$b$}
\qline(2.25,0.25)(4,0.25)
\psdot(3,1.5)
\put(3,1.55){$c$}
\psdot(2.75,0.25)
\put(2.75,0.14){$p$}
\qline(3,1.5)(2.25,0.25)
\qline(3,1.5)(4,0.25)
\pscircle(2.877,0.88){0.03}
\put(2.92,0.92){$q$}
\qline(3,1.5)(2.88,0.9)
\qline(2.75,0.25)(2.787,0.432)
\qline(2.876,0.86)(2.795,0.492)
\psdot(2.65,0.915)
\put(2.53,0.9){$s$}
\psdot(3.58,0.775)
\put(3.63,0.76){$t$}
\qline(2.65,0.915)(2.85,0.885)
\qline(3.57,0.775)(2.91,0.875)
\psline[linestyle=dashed](2.25,0.25)(2.76,0.453)
\psline[linestyle=dashed](3.58,0.775)(2.82,0.47)
\pscircle(2.79,0.462){0.03}
\put(2.81,0.365){$r$}
\endpspicture}

\begin{document}
\maketitle

\begin{abstract}
We report on a project to use a theorem prover 
to find proofs of the theorems in 
Tarskian geometry.
These theorems start with fundamental properties of betweenness, proceed
through the derivations of several famous theorems due to Gupta 
and end with the derivation from Tarski's axioms of Hilbert's 1899 axioms for geometry.
They include the four challenge problems left unsolved by Quaife, who two decades ago
found some \Otter proofs in Tarskian geometry (solving challenges issued in Wos's 1998 book).

There are 212 theorems in this collection.  We were able to 
find \Otter proofs of all these theorems.  We developed 
a methodology for the automated preparation and checking of the input files for 
those theorems, to ensure that no human error has corrupted the formal development of an entire theory
as embodied in two hundred input files and proofs. 

We distinguish between proofs that were found completely mechanically (without 
reference to the steps of a book proof)  and proofs that were constructed by 
some technique that involved a human knowing the steps of a book proof. 
Proofs of length 40--100, roughly speaking, are difficult
exercises for a human,  and proofs of 100-250 steps belong in a Ph.D. thesis or
publication.  29 of the proofs in our collection are longer than 40 steps,
and ten are longer than 90 steps.  
We were able to derive completely mechanically all but 26 of the 183 theorems that have 
``short'' proofs (40 or fewer deduction steps).  We   found proofs of the 
rest, as well as the 29 ``hard'' theorems,
  using a method that requires consulting the book proof at the outset.
Our ``subformula strategy'' enabled us to prove four of the 29 hard theorems
completely mechanically.  These are Ph.D. level proofs, of 
length up to 108.   
 
\end{abstract}.



\section{Introduction}

 Geometry has been a testbed for automated deduction almost
as long as computers have existed; the first experiments were done
in the 1950s.  In the nineteenth century, geometry was the testbed for the development 
of the axiomatic method in mathematics, spurred by the efforts to prove Euclid's
parallel postulate from his other postulates and ultimately the development of
non-Euclidean geometry.  This effort culminated in Hilbert's seminal 1899 book \cite{hilbert1899}.
 In the period 1927--1965, Tarski developed his 
simple and short axiom system (described in \S\ref{section:axioms}). 
Some 35 years ago, Wos experimented with finding proofs
from Tarski's axioms, reporting success with simple theorems, but leaving several
unsolved challenge problems.   The subject was revisited
by Art Quaife, who in his 1992 book \cite{quaife1992} reported on the successful 
solution of some of those challenge problems using an early version of McCune's theorem prover, \Ott.  
But  several theorems remained that Quaife was not able to get \Otter to prove,
and he stated them as ``challenge problems'' in his book.
As far as we know, nobody took up the subject again until 2012, when we 
set out to see whether automated reasoning techniques, and/or computer
hardware,  had improved enough to let us progress beyond Quaife's achievements.%
\footnote{There is also a long tradition, going back to Descartes, of reducing geometric
problems to algebra calculations by introducing coordinates.  Algorithms for carrying out 
such calculations by computer have been extensively studied, including special  methods intended
for geometry and 
Tarski's general decision  procedure for real closed fields.  We mention these only to 
emphasize that such methods are irrelevant to this paper, which is concerned with proofs
in an axiomatic system for geometry.}

The immediate stimulus leading to
 our 2012 work was the existence of the almost-formal development of 
many theorems in Tarskian geometry in Part I of \cite{schwabhauser}.  This Part I 
is essentially the manuscript developed by Wanda Szmielew for her 1965 Berkeley lectures
on the foundations of geometry, with ``inessential modifications'' by Schw\"abhauser.
There are 16 chapters.  Quaife's challenge problems (listed in \S\ref{section:challenges})
 occur in the first nine chapters.
The rest contain other important geometrical theorems (described in \S\ref{section:results} 
and \S\ref{section:hardtheorems}).
We set ourselves the goal to find \Otter proofs of each of the theorems in
Szmielew's 16 chapters; we completed twelve of them, which is enough to make our points.
Our methodology focuses on separate problems, each targeting one theorem.  For
each problem, we supplied the axioms and the previously-proved 
theorems, as well as the (negated) goal expressing the theorem to be proved.
We were often forced to supply \Otter with more information than that,
as we describe in \S\ref{section:diagrams} and \S\ref{section:hints}.

We do not know of another 
body of mathematics of this size that has been formalized by using a theorem prover. 
Normally proof checkers are used to formalize a whole theory, and theorem provers
are used for individual theorems.  Our first report on this work \cite{beeson2014-wos}
emphasized the solution of two hundred individual problems.  We prepared two 
hundred \Otter input files by hand.   Since the time of that report, we developed 
a methodology for the automated preparation and checking of those input files, to 
ensure that no human error has corrupted the formal development of an entire theory
as embodied in two hundred input files and proofs.   This methodology makes it 
possible to easily conduct experiments involving many files.  For example,  
one can easily generate input files for the theorem prover of one's choice, 
instead of for the prover we used (\Ott).  
On the web site for this project \cite{tarski-archive}, we have made available
our input files, the resulting proofs, and various tools for manipulating theorems
and files,  which we hope will be of use to others wishing to work with these problems.

We distinguish between proofs that were found completely mechanically (without 
reference to the steps of a book proof)  and proofs that were constructed by 
some technique that involved a human knowing the steps of a book proof.  Roughly 
speaking, we were able to derive mechanically most of the theorems that have 
``short'' proofs (40 or fewer deduction steps).     
Proofs of length 40--100, roughly speaking, are difficult
exercises for a human,  and proofs of 100-250 steps belong in a Ph.D. thesis or
publication.  At first (that is, in \cite{beeson2014-wos}),  we could not obtain
such long proofs completely mechanically.   Our main goal at that time (2013)
was to obtain formal proofs by any means possible.  That meant starting with 
a book proof  and using techniques explained in \S\ref{section:hints} (lemma adjunction and hint injection)
to eventually obtain a formal proof.  We did obtain proofs of all two hundred theorems.  
But that left an obvious challenge:  Can one develop methods that 
enable a theorem prover to prove these theorems without reference to the steps of a book proof?
We were able to prove some (but not all) 
quite long theorems (of Ph.D. difficulty) 
completely mechanically,  using a technique called the ``subformula strategy''
\cite{wos-notebook2008}.
The challenge of developing strategies to reliably find proofs 
of such difficult theorems (by methods independent of knowledge of a book proof) still stands.  

In this paper, we give Tarski's axioms, explain the challenge problems of Quaife
and some of the axioms of Hilbert, discuss the difficulties of finding \Otter 
proofs of these theorems, and explain what techniques we used to find those proofs. 
Then we explain our automated file-generation
and checking methodologies and the extent to which those methods ensure the reliability
of the whole theory.

\section{The challenge problems} \label{section:challenges}
Quaife's four challenge problems were as follows: every line segment has a midpoint;
every segment is the base of some isosceles triangle; the outer Pasch axiom (assuming 
inner Pasch as an axiom); and the first outer connectivity property of betweenness.
These are to be proved without any parallel axiom and without even line-circle continuity.
The first proofs of these theorems were the heart of Gupta's 
Ph.D. thesis \cite{gupta1965}  under Tarski.  \Otter proved them all in 2013.   
All Quaife's challenge problems occur in Szmielew; the last of them is Satz 9.6,
so solving these challenge problems is a consequence of formalizing Chapters 2-11 of 
Szmielew.   

The theory of Hilbert (1899) can be translated into Tarski's language, interpreting
lines as pairs of distinct points and interpreting angles as ordered triples of non-collinear points.
Under this interpretation, the axioms of Hilbert either occur among or are easily 
deduced from theorems in the first 11 (of 16) chapters of Szmielew.  We have found
\Otter proofs of all of Hilbert's axioms from Tarski's axioms (through Satz 11.49 
of Szmielew, plus Satz 12.11).

\section{Related work}
We know of several other projects involving formal proofs in Tarskian geometry.
Braun and Narboux \cite{narboux2012}, \cite{narboux2015} have  checked many theorems of Tarskian geometry in Coq; they have now gotten as far as Pappus's theorem in Chapter 15. 
 Richert has checked some in HOL Light (unpublished, but distributed with HOL Light).  
 Urban and Veroff are also checking some in HOL-Light.  
 Durdevic {\em et.al.} \cite{narboux2015b} have used Vampire, E, and {\textsc SPASS}
 to check some of these theorems.   
 
Wos has also conducted many experiments aimed at shortening some of 
these proofs, and other proofs in Tarski's older axiom system,  or finding forward 
proofs instead of backwards or bidirectional proofs;  some of these are
described in \cite{wos-notebookTarski}.

\section{Tarski's axioms} \label{section:axioms}
 In about 1927,  Tarski first lectured on his axiom system for geometry, which 
was an improvement on Hilbert's 1899 axioms in several ways: First, the language
had only one sort of variables (for points), instead of having three primitive
notions (point, line, and angle).   Second, it was a first-order theory (Hilbert's
axioms mentioned sets, though not in an essential way).   Third, the axioms were short,
elegant, and few in number (only twenty in 1927, decreasing to twelve in 1957 \cite{tarski-givant}).  They could be expressed comprehensibly
in the  primitive syntax, without abbreviations.   

\subsection{History}
 The development of Tarski's theory, started in 1927 or before, was delayed, first 
 by Tarski's involvement in other projects, and then by  
World War II (galley proofs of Tarski's article about it were destroyed by bombs).
The first publication of Tarski's axioms came in 1948 \cite{tarski1951} and contained little more 
than a list of the axioms and a statement of the important metamathematical theorems 
about the theory (completeness, representation of models as $\F^2$ for $\F$ a 
real-closed field, quantifier-elimination, and decidability).  Tarksi
then lectured on the subject at Berkeley in 1956 to 1958, and published a reduced
set of axioms in 1959 \cite{tarski1959}.  In the 1960s, 
Tarski, Szmielew, Gupta, and Schw\"abhauser (and some students) 
reduced the number of axioms still further.  
The manuscript that Szmielew prepared for her 1965 course became Part I of \cite{schwabhauser}. 
More details of the history of these axioms can be found in \cite{tarski-givant} (our main source) 
and 
the foreword to (the Ishi Press edition of) \cite{schwabhauser}. 
For our present purposes, the relevance of the
history is mainly that there are three versions of Tarski's theory: the 1948 version, 
the 1959 version, and the 1965 version (published in 1983).
The earlier experiments of Wos used the 1959 axioms, but Quaife used the 1965 version,
as we do.   The exact differences are explained in \S\ref{section:betweenness} and 
\S\ref{section:pasch}. 

\subsection{Syntax}
The fundamental relations in the 
theory (first introduced by Pasch in 1852) are ``betweenness'',  which we here write
$\T(a,b,c)$,  and ``equidistance'', or ``segment congruence'',  which is officially 
written $E(a,b,c,d)$  and unofficially as $ab = cd$, segment $ab$ is congruent to segment
$cd$.  The intuitive meaning of $\T(a,b,c)$ is that $b$ lies between $a$ and $c$ on
the line connecting $a$ and $c$;  Tarski used non-strict betweenness, so we do 
have $\T(a,a,c)$ and $\T(a,c,c)$ and even $\T(a,a,a)$. Hilbert used strict betweenness.
Both of them  wrote $\B(a,b,c)$, which is a potential source of confusion.  We therefore
reserve $\B$ for strict betweenness and use $\T$ for Tarski's non-strict betweenness.
The fact that Tarski's initial is `T' should serve as a mnemonic device.  Of course
the equality relation between points is also part of the language.  

\subsection{Betweenness and congruence axioms}\label{section:betweenness}
We sometimes write $ab=cd$ instead of $E(a,b,c,d)$ to enhance human readability. In \Otter files
of course we use $E(a,b,c,d)$.  The following are five axioms from the 1965 system.
\smallskip

\axioms
$ab = ba$& (A1)  reflexivity for equidistance\\
$ab=pq \land ab = rs \implies pq = rs$& (A2) transitivity for equidistance \\
$ab = cc \implies a=b$& (A3) identity for equidistance \\
$\exists x\,(\T(q,a,x) \land ax = bc)$& (A4) segment extension\\
$\T(a,b,a) \implies a=b$& (A6) identity for betweenness
\endaxioms

When using (A4) in \Ott, we Skolemize it:
\smallskip

\axioms
$ \T(q,a,ext(q,a,b,c)) \land E(a,ext(q,a,b,c),b,c)$& (A4) Skolemized\\
\endaxioms

The original (1948) theory had the following additional fundamental properties
of betweenness listed as axioms. (We follow the numbering of \cite{tarski-givant}.)
\smallskip

\hskip-0.5cm
\axioms
$\T(a,b,b)$&  (A12) Reflexivity for $\T$ \\
$\T(a,b,c) \implies \T(c,b,a)$&  (A14) Symmetry for $\T$   \\
$\T(a,b,d) \land \T(b,c,d) \implies \T(a,b,c)$& (A15) Inner transitivity \\
$\T(a,b,c) \land \T(b,c,d) \land b \neq c \implies \T(a,b,d)$&(A16) Outer transitivity \\
$\T(a,b,d) \land \T(a,c,d) \implies \T(a,b,c) \lor \T(a,c,b)$ & (A17) Inner connectivity \\
$\T(a,b,c) \land \T(a,b,d) \land a\neq b$& (A18) Outer connectivity \\
$\qquad \implies \T(a,c,d) \lor \T(a,d,c)$& 
\endaxioms
\smallskip

\noindent
Of these only (A15) and (A18) appear in the 1959 version, because in 1956--57 Tarski and 
his students Kallin and Taylor showed that the other four are dependent (derivable from the 
remaining axioms).  H.~N.~Gupta
showed in his 1965 Ph.D. thesis \cite{gupta1965} that (A18) is also dependent.  The 
proof of (A18) is one of Quaife's challenge problems.
Gupta also showed that (A15) implies (A6) using the other axioms of the 1959 system.
Then one could have dropped (A6) as an axiom; but instead, Szmielew
dropped (A15), keeping (A6) instead; then (A15) becomes a 
theorem.   

All six of these axioms  occur
as theorems in \cite{schwabhauser}:  (A12) is Satz~3.1, (A14) is Satz~3.2, (A15) is Satz~3.5,
(A16) is Satz~3.7, (A18) is Satz~5.1, and (A17) is Satz~5.3.  
Hence, our research program of proving all the 
theorems in Szmielew's development using \Otter  systematically captured these results as 
soon as we reached Satz~5.3. 

\subsection{The five-segment axiom}
\begin{figure}[ht]
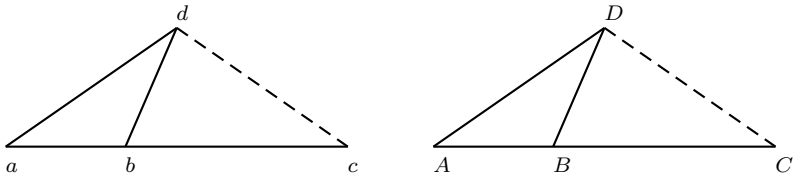

\TarskiFiveSegmentFigure
\caption{The five-segment axiom (A5)}
\label{figure:5segment}
\end{figure}

Hilbert \cite{hilbert1899} treated angles as primitive objects and angle congruence
as a primitive relation, and he took SAS (the side-angle-side triangle congruence principle) 
as an axiom. In Tarski's theory, angles are
  treated as  ordered triples 
of points, and angle congruence is a defined notion,
so a points-only formulation of the SAS principle is required.   
 The key idea is Tarski's  ``five-segment axiom'' (A5), shown in Fig.~\ref{figure:5segment}.

If the four solid segments in Fig.~\ref{figure:5segment} are pairwise congruent, then the fifth (dotted) segments are congruent too.
This is essentially SAS for triangles $dbc$ and $DBC$.  The triangles $abd$ 
and $ABD$ are surrogates, used to express the congruence of angles $dbc$ and $DBC$.
 By using Axiom A5, we 
can avoid all mention of angles.

\subsection{Pasch's axiom} \label{section:pasch}
\begin{figure}[ht]
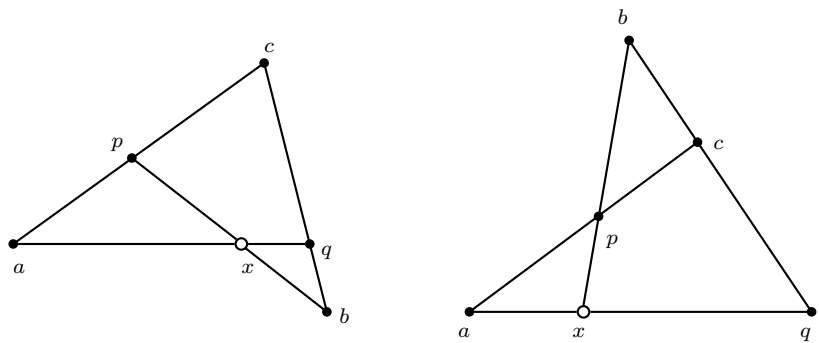
   
\hskip 0.7cm
\InnerOuterPaschFigure
\caption{ Inner Pasch (left) and Outer Pasch (right).  Line $pb$ meets triangle $acq$ in one side.
 The open circles show the points asserted to exist on the other side. }
\label{figure:InnerOuterPaschFigure}
\end{figure}

Moritz Pasch \cite{pasch1882} (see also \cite{pasch1926}, with an historical appendix by Max Dehn)
 supplied (in 1882) an axiom that repaired many of the defects that 
nineteenth-century rigor found in Euclid.  Roughly, a line that enters a 
triangle must exit that triangle.
As Pasch formulated it,  it is not in $\forall\exists$ form.  There are two $\forall\exists$  versions,
illustrated in Fig.~\ref{figure:InnerOuterPaschFigure}.  These formulations of Pasch's axiom 
go back to Veblen \cite{veblen1904}, who proved outer Pasch implies inner Pasch. 
Tarski took outer Pasch as an axiom in \cite{tarski1959}.

\smallskip

\axioms
$\T(a,p,c) \land \T(b,q,c) \implies \exists x\,(\T(p,x,b) \land \T(q,x,a))$& (A7) inner Pasch \\
$\T(a,p,c)\land \T(q,c,b) \implies \exists x\,(\T(a,x,q) \land \T(b,p,x))$&  outer Pasch 
\endaxioms
\smallskip

For use in \Ott, we introduce Skolem symbols 
$$ip(a,p,c,b,q) \qquad \mbox{ and} \qquad   op(p,a,b,c,q)$$ 
for the point $x$ asserted to exist.  The use of these symbols makes these axioms
quantifier free.

Tarski originally took outer Pasch as an axiom.  In \cite{gupta1965}, Gupta proved
both that inner Pasch implies outer Pasch and that outer Pasch implies inner Pasch,
using the other axioms of the 1959 system.   
The proof of outer Pasch from inner Pasch
is one of Quaife's four challenge problems.
 
 \subsection{Dimension axioms}
With no dimension axioms, Tarski's geometry axiomatizes theorems that are true in $n$-dimensional
geometries for all $n$.  For each positive integer $n$, we can specify that the dimension of 
space is at least $n$ (with a lower-dimension axiom A8${}^{(n)}$), or at most $n$ (with an 
upper-dimension axiom A9${}^{(n)}$).  The uppe--dimension axiom says (in a first-order way) that the set of points
equidistant from $n$ given points is at most a line.   The lowe--dimension axiom for $n$ is the 
negation of the upper-dimension axiom for $n-1$.  For the exact statements of these axioms 
see \cite{tarski-givant}.

Inner and outer Pasch have another advantage over Pasch's original version, besides 
logical simplicity.  Namely, they hold even in 3-space, where Pasch's original version fails.
That is, they can be used without an ``upper-dimension'' axiom, whose use is thereby postponed
 a long time in Szmielew.

\subsection{Tarski's parallel axiom (A10)}
\begin{figure}[ht]
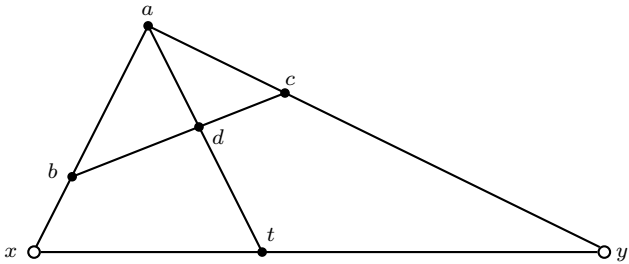

 \hskip 2cm
 \TarskiParallelFigure 
 \caption{Tarski's parallel axiom}
 \label{figure:parallelaxiom}
 \vskip -1.2cm
 \end{figure}

In  Fig.~\ref{figure:parallelaxiom}, open circles indicate points asserted to exist.
 \vskip-0.5cm

\begin{eqnarray*}
& \T(a,d,t) \land \T(b,d,c) \land a \neq d \implies  \\
& \qquad \exists x \exists y\,(\T(a,b,x) \land \T(a,c,y) \land \T(x,t,y)) \qquad ({\rm A}10)  
 \end{eqnarray*}

 The hypothesis says that $t$ lies in the interior of angle $a$, as witnessed by $b$, $c$, and $d$.
 The conclusion says that some line through $t$ meets both sides of the angle.  Of course this fails
 in non-Euclidean geometry when both $ab$ and $ac$ are parallel to some line $L$ through $t$.
 
  According to \cite{tarski-givant}, Szmielew preferred to use the ``triangle circumscription 
  principle '' (A10${}_2$) as the parallel axiom.  Substituting (A10${}_2$) was apparently one of the 
``inessential changes'' made by Schwabh\"auser.  This principle says that if $a$, $b$, and
 $c$ are not collinear, then there exists a point equidistant from all three. (That point is the 
 center of the circumscribed triangle, hence the name.)  See \cite{tarski-givant}
 for a formal statement.
 
 \subsection{Continuity axioms}
 Axiom schema (A11) is not a single axiom but an axiom schema, essentially asserting that 
 first-order Dedekind cuts are filled.  See \cite{tarski-givant}
 for a formal statement.
 Models of A1-A11 are all isomorphic to planes $\F^2$
 where $\F$ is a real-closed field.  One can also consider instead of (A11) the axioms of 
 line-circle continuity and/or circle-circle continuity,  which assert the existence of intersection 
 points of lines and circles, or circles and circles, under appropriate hypotheses.  None of 
 the continuity axioms are used in the work reported in this paper.  Szmielew's development 
 proceeds strictly on the basis of A1-A10.

\section{Completely mechanical methods for finding proofs}
In this section we describe, with illustrative examples, the techniques we used 
that do not involve knowing the steps of a book proof (not 
even the diagram accompanying the proof).
 
\subsection{How \Otter works}
Readers familiar with \Otter can skip this subsection.  It is not an introduction to \Otter 
but an attempt to make the subsequent information about our methodology comprehensible to 
those who do not have expertise with \Ott; at least, it should enable such readers to 
make sense of the input files and proofs we exhibit on the project's website
\cite{tarski-archive}.
For more information about \Ott, see \cite{wos-fascinating}.

\Otter is a clause-based resolution theorem prover.  One writes {\tt -A}  for the 
negation of $A$.   One writes {\tt A | B} for disjunction (``or'').  One does not 
write ``and'' at all, but instead one enters the two clauses separately.  One writes
$A \implies B$  as {\tt -A | B}.   Similarly one writes $P \land Q \implies R$ as 
{\tt -P | -Q | R}. 

Variables begin with {\tt x,y,z,w,u,v}.  Names beginning with any other letter are constants.
A resolution theorem prover requires the goal to be negated and entered as clauses.
For example, to prove $A(x) \implies \exists y\, B(x,y)$,  we would enter the following clauses:

\begin{verbatim}
A(c).
-B(c,y).
\end{verbatim}
After proving this theorem, if we want to use it to prove the next theorem, we invent 
a new Skolem symbol {\tt f} and enter the theorem as
{\tt -A(x) | B(x,f(x))}.

The input to \Otter is contained in two lists,  the ``set of support'' (sos) and 
``usable''.  The fundamental run-time loop of \Otter moves a clause from sos to usable,
and then tries to use one of the specified inference rules to generate new clauses from 
that clause and other clauses in usable.  If conclusions are generated, \Otter has to decide
whether to keep them.  If it decides to keep them, they are placed on sos, where they
can eventually be used to generate yet more new clauses.   If the empty clause is generated,
that means a proof has been found, and it will be output.  

The fundamental problem of automated deduction is to avoid drowning in a sea of useless conclusions
before finding the desired proof.  One tries to get control over this by assigning ``weights''
to clauses, adjusting those weights in various ways, and using them to control both which 
clauses are kept and which clause is selected from sos for the next iteration of the loop.
By default: the weight of a clause is the number of its symbols; the next clause selected
is the lightest one in sos; and clauses are kept if their weight does not exceed a 
parameter {\tt max\_weight}.   More sophisticated ways of setting the weights have been developed
over the past decades and are discussed in \S\ref{section:hints}.
  The idea is to get the weights of the 
important clauses to be small, and then to squeeze down {\tt max\_weight} to prevent
drowning.

In addition to techniques involving weighting, there are other ways to control \Ott's search:
\begin{itemize}
\item Use a propitious combination of rules of inference.  For an introduction to these rules please
refer to \cite{wos-fascinating}. 
\item One can exert some control over which clause will
be selected from sos at the next iteration by using \Ott's {\tt pick\_given\_ratio}.  
\item One can exert some control over how the search starts and what kind of proof to seek
(forward, backward, or bi-directional) by choosing which clauses to put in sos and which 
to put in usable.
\end{itemize}

\subsection{Hints}  
Putting a clause 
into {\tt list(hints)} causes \Otter to give that clause, if deduced, a low weight, causing 
it to be retained, even if its default weight would have been so large as to cause it to be 
discarded.   One has options (specified at the top of an \Otter file) to cause this weight 
adjustment to apply to clauses that match the hints, or subsume the hints, or are 
subsumed by the hints.  We use hints in the implementation of several strategies, 
including the subformula strategy and lemma adjunction, which are described 
in \S\ref{section:subformulastrategy} and \S\ref{section:hints}.
The technique of hints was invented by Veroff \cite{veroff1996} and later incorporated
into \Ott.  As a technical note: when using hints, one should always include these lines,
without which the hints will not have the desired effect.

\begin{verbatim}
assign(bsub_hint_wt,-1).
set(keep_hint_subsumers).
\end{verbatim}

There is also a technical difference in \Otter between {\tt list(hints)} and {\tt list(hints2)};
 researchers should consult the manual before using either.  
Another similar technique is known as {\em resonators}.  This is more useful when one has
a proof in hand and wishes to find a shorter proof.  For the exact differences between 
hints and resonators, see \cite{wos2003}, p.~259.

\subsection{Choice of inference rules and settings}
We mentioned that one of the ways \Otter can be controlled is through a 
propitious choice of inference rules and settings.  We tinkered with our choices often,
in the hope that a different choice would be better.  We did not find that one 
choice was always best, but a carefully chosen default choice served us well 
in most cases.  Our default choice was to use all three of 
hyperresolution, binary resolution, and unit resolution.  In addition, 
we always used paramodulation (for equality reasoning). 
   We {\em often} 
changed the values of the parameters {\tt max\_weight} and {\tt max\_proofs}.
In particular, when using lemma adjunction, we needed {\tt max\_proofs} large enough 
to accommodate many intermediate goals (on the passive list); when using many hints,
we chose {\tt max\_weight} to be 8 or even smaller;  but when searching a ``raw''
search space, we often used 16, 20, or 24, and occasionally even higher. 
If {\tt max\_weight} is too high, then one may be swamped with too many conclusions,
but if {\tt max\_weight} is too low, one may throw away a clause that was needed.  Novices
may fail to realize another danger:  one might derive a useful clause,  and keep it, 
but if, say, its weight is 10 and there are many clauses of smaller weight, your 
useful clause may ``never'' become the given clause and give rise to the further 
conclusions that are needed to complete the proof.

Occasionally we changed the values of 
{\tt max\_distinct\_vars},  but usually without much effect.  Rarely do the 
proofs we are seeking  contain more than one variable per clause.  We think 
this is because Szmielew has already organized the theorems in such a way that we 
do not need to prove many extra lemmas;  a derived clause with variables amounts to a lemma.
We performed experiments in which {\tt max\_distinct\_vars} was given increasing values,
holding other settings fixed.  There were only about four theorems that required 
it to have the value 4,  but setting it lower did not speed up finding the other proofs
noticeably.

We used paramodulation for equality reasoning.  Wos believes that
failing to use paramodulation was an important reason for the limited success
of the experiments he made in the 1980s in Tarskian geometry; but we also note
that Quaife writes that paramodulation was available for use but seldom actually used in his proofs.

\paragraph{Demodulation.}  A {\em demodulator} is an equation, used left-to-right 
to rewrite terms to ``simpler forms.''  We used demodulators in connection with 
diagram equations, but usually we could also obtain proofs without those demodulators.
In other subjects, Wos has made use of the technique of ``demodulating to junk'',
in which unwanted classes of term or types of conclusions are demodulated to an atom {\tt junk},
which is eliminated by giving it a high weight.  But this technique was not useful in 
our work on geometry.  Only one of the two hundred theorems of Tarskian geometry 
is an equation.  That theorem says that reflection in a point is an involution; that is, 
the reflection of the reflection of a point is the original point.  We write $s(p,x)$ for 
the reflection of $x$ in $p$;  then the equation is $s(p,x(p,x)) = x$.%
\footnote{There is a similar theorem, Satz 10.5, about reflection in a line, but that
theorem is not quite equational, because it requires that the two points determining the 
line be distinct.}     It was 
 useful to make this equation a demodulator;  but since there is  only {\em one}
equational theorem, demodulation is not an important technique in the development
of Tarskian geometry based on Szmielew's manuscript.

We did, however, make some use of demodulators in the process of lemma adjunction.
For example, $midpoint(x,y) = midpoint(y,x)$ is a useful equation (which does not occur
in Szmielew).   Also, in several situations   it is natural to rewrite 
(not terms but) formulas, for example, we would naturally regard all the following 
as equivalent ways to express the congruence of two segments:  $ab = cd$, $ab=dc$,
$ba = cd$, $ba = dc$,  $cd = ab$, $dc=ab$, $cd = ba$, and $dc = ba$.  Similarly 
for the 4-ary predicate $ab \perp cd$.   \Otter is able to use demodulators for 
formulas, and setting the {\tt lrpo} flag tells it to use them to demodulate terms 
to what amounts to alphabetical order; thus we would keep $midpoint(a,b)$ unchanged
and demodulate $midpoint(b,a)$ to $midpoint(a,b)$.  Technically, however,  the use of 
demodulators at a formula level is not permitted in a first-order development, so 
in the end we tried (successfully) to eliminate their use, mostly by hand-editing the 
hints resulting from proof steps using demodulators.  Like Gauss, we removed our 
scaffolding,  so this paragraph is the only trace of that technique.

\paragraph{  Hot list.}  The hot list is a feature of \Otter\ that works as 
follows:  Whatever clauses are on the hot list are applied immediately to newly deduced
formulas.  This is useful in the following situation:  suppose you think that clause
$P$ should be derived, and then clause $Q$ should be derived from $P$ by axiom 7. 
You see that $P$ is derived, but then $P$ never becomes the given clause,  because 
it has a fairly high weight and the sea of conclusions contains many clauses of 
lower weight.   If you put Axiom 7 on the hot list, then $Q$ will be deduced immediately
when $P$ is deduced, without having to wait until $P$ is the given clause.
We used this technique in perhaps fifty out of two hundred hand-crafted input 
files.   Perhaps it wasn't really necessary.  In the mechanically generated
files described in \S\ref{section:mechanicalinputfiles}, the hot list disappears, but its influence lingers, because
the steps of the proof found originally using a hot list remain as hints. 
 
\paragraph{Set of support.} Normally, we placed the axioms and previously proved theorems
on the list of ``usable'' clauses, and the negated form of the current theorem on the 
set of support.  One of the things that can be tried is putting 
everything  on the set of support.  This goes against the basic idea behind the set of support (to
prevent deducing many consequences of the axioms that are irrelevant to the current goal),
but it has worked for Wos in some situations.  In geometry, however, it 
worked on exactly one theorem out of two hundred, namely Satz 3.1.

\paragraph{ Level saturation.}  ``Level saturation''
 refers to a gradual increase in allowed complexity.
The program starts
by allowing the simplest next conclusions, for us of weight 1, then 2, 3, $\ldots$,
 and continues, run after run.   This will be useful only if there is a very short proof
 that for obscure reasons does not come out with more normal settings.   Once or twice
 we found a short proof this way, but we always found a proof later with more normal settings,
so this was not actually a useful technique.

\paragraph{Ratio strategy.}  The pick-given ratio in \Otter alters the basic strategy 
of selecting the lowest-weight clause from the set of support.  For example, if this 
ratio is set to 4 (our default), then after four clauses have been selected by weight,
the oldest one on the set of the support is chosen, regardless of weight.  We sometimes,
but rarely, had success by setting this ratio to 2.

\section{Using the diagram} \label{section:diagrams}
In this section, we consider the method of ``diagrams'',  which involves looking at the diagram 
in the book accompanying the proof,  but not the steps of the proof itself.

We learned this technique from 
Quaife \cite{quaife1992}.  Suppose we are trying to find an $x$ such that 
$A(a,b,x)$.   If we let a theorem prover search, without making any attempt to guide the search, 
essentially it will generate all the points that can be constructed from $a$ and $b$ (and 
other points in the problem or previously constructed in clauses that were kept)
by the Skolem functions for segment extension and inner Pasch.  Imagine trying to prove 
a geometry theorem that way:  just take your ruler, draw all the lines you can between 
the points of the diagram,  label all the new points formed by their intersections (that is
by using 
Pasch), and construct every point that can be constructed by extending a segment you have by 
another segment you have.  See if any of the new points is the point 
you are trying to construct. If not,  repeat.  You will generate a sea of useless points,
even if you discard those with too many construction steps.  

\begin{figure}[ht]
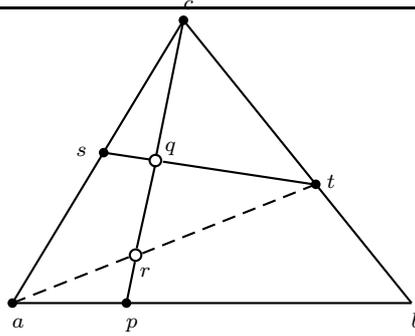

\hskip 0.3in
\begin{verbatim}
     list(sos).
     T(a,s,c).
     T(b,t,c).
     T(a,p,b).
     -T(p,x,c)|-T(s,x,t).
     r = ip(c,t,b,a,p).
     q = ip(c,s,a,t,r).
     end_of_list.
\end{verbatim}
\vskip-1.6in
\CrossBarFigure
\caption{The crossbar theorem asserts that $q$ exists, given the other points
except $r$.
To prove it, we construct first $r$ and then $q$, using inner Pasch twice. Here {\tt ip}
is the Skolem function for the inner Pasch axiom.}
\label{figure:crossbar}
 
\end{figure}

To guide \Otter (or any other theorem prover)
 to construct the right points, we ``give it the diagram'' by 
defining the points to be constructed, using  Skolem functions.  For example, 
consider the ``crossbar theorem'' (Satz 3.17).  See Figure~\ref{figure:crossbar},
 shows the diagram and the input (not shown are the axioms A1-A6 
and the previous theorems, which are placed in {\tt list(usable)}).
The two lines defining $r$ and $q$ are what we call ``giving \Otter the diagram.''
With those lines present, \Otter finds a proof instantly.  Remove them, 
and \Otter does not find a proof (at least, not in ten minutes).

The reason this works is that $q$, being a single letter, has weight 1,  so 
clauses involving $q$ have much lower weight than would the same clauses with 
$q$ replaced by $ip(c,s,a,t,r)$, which has weight 6.   A clause involving the 
terms defining both $q$ and $r$ would have weight at least 14,  so it would 
not be considered very soon, and meantime many other points will be constructed.

In this simple example, if one puts all clauses into sos and nothing in usable, 
\Otter does eventually find a proof without being given the diagram; but it is certainly 
much slower than with the diagram.  In more complex examples, we  
think this technique is essential.  Here, for example, are the lines 
from the input file for Satz~5.1, the inner connectivity of betweenness, 
describing the two  complicated diagrams on pp.~39--40 of \cite{schwabhauser}:

\begin{verbatim}
         c1=ext(a,d,c,d).  
         d1=ext(a,c,c,d).
         p=ext(c1,c,c,d).
         r=ext(d1,c,c,e).
         q=ext(p,r,r,p).
         b1 = ext(a,c1,c,b).
         b2 = ext(a,d1,d,b).
         e = ip(c1,d,b,d1,c).
\end{verbatim}
 
\section{The subformula strategy} \label{section:subformulastrategy}
Wos invented the subformula strategy sometime in 2007--2008 and 
used it to study single axioms in relevance logics.  He wrote up this work 
in \cite{wos-notebook2008}.  
  The subformula strategy consists of this:

\begin{quote}
{\em For every literal occurring in a clause in the set of support,
put both the literal and its negation into the hints list.}
\end{quote}

\noindent
Since clauses that subsume a hint are kept and given low weight, the result is that 
when any subformula of a clause in the set of support is deduced, it is kept, and 
moreover,  it will soon be used as the ``given clause'' to deduce more conclusions.
The idea is that the proof we are seeking is likely to involve such clauses, so 
when we deduce them, we are getting closer to a good result.

This strategy is particular easy to implement in a modern programming language 
with a string manipulation function like PHP's {\tt explode} or Python's {\tt split}.
No parsing is required.  Just explode each clause in the set of support on the 
symbol for disjunction, resulting in an array of literals. Negating a literal is 
also a simple string manipulation: for example, in PHP one can use {\tt str\_replace} to 
replace {\tt !=} by {\tt =} and vice-versa.  To compute the required hints and write 
them into an already-generated input file takes less than 30 lines of code.

This strategy is completely general;  it can be applied to any problem and 
requires no advance knowledge of a proof.  Perhaps not even its creator 
noticed its generality at the time.  
 We failed to use
the subformula strategy in 2012--2014,  and at that time we found no proofs of length 
greater than 30 in a completely mechanical fashion, that is, without any reference to 
a book proof  (although we did find proofs of all the Tarski theorems we tried,
by methods discussed in \S\ref{section:hints} that did start by reference to a book proof).

When the light bulb came on over our heads, we used the subformula strategy  
and found several amazingly long proofs completely mechanically,  as described in 
\S\ref{section:results} and \S\ref{section:hardtheorems}.  We kept {\tt max\_weight}
as low as possible,  typically 8.

\section{Methods that require using steps of the book proof} \label{section:hints}
In this section we describe a method, or perhaps combination of methods, 
that starts with the steps of a book proof (thus, human steps rather than 
machine steps)   and ends with an \Otter proof.

Our methodology was as follows:  when trying to prove a theorem, we prepared an 
input file with the negated goal in {\tt list(sos)}  and the axioms and previously proved theorems 
in list(usable), and a default choice of inference rules.  If this 
did not find a proof, and the proof in Szmielew had a diagram, we supplied the diagram.
This section concerns what we did with theorems that could not be proved 
from these input files alone.  Namely, we supplied some hints based on the book proof.
Here are the details about how this idea led to proofs.

\paragraph{From the passive list to hints.}
We tried supplying some intermediate goals, namely, 
some or even all of the steps of the book proof.

These steps were supplied in two places:  we put them in the hints list, 
and we also put their negated forms in {\tt list(passive)},
with answer literals having numbers for those goals.  The passive list is used
as follows:  When a clause is generated that conflicts with an element of the passive
list, a proof is generated.  After that, proof search continues.
Hence, when this file is run,
one sees which of the intermediate goals are being proved.  Among other things, this 
showed us where \Otter was ``getting stuck.''   But even though we found no proof of the 
main goal, we had proofs of some intermediate goals.  We converted these proof steps
to hints  and ran \Otter again.   Sometimes more intermediate goals would be reached. 
One can tinker with {\tt max\_weight}: sometimes a smaller {\tt max\_weight} may keep 
one from drowning, and sometimes a larger {\tt max\_weight} may allow a vital clause 
to be kept.  
After adding the new hints, the previous proofs were found immediately, and perhaps
more intermediate goals could be found,  resulting in still more hints.

With luck this process would converge to a proof.

\paragraph{Lemma adjunction and hint injection.}  Often, especially with more difficult theorems,
the process described above would lead not to a proof but to a situation in which 
some of the book steps would not be proved, even though others would be.  In that
case, our next move was to place one or more of these steps,  or their negations, 
in the set of support.  If this generated a proof (of the original goal, or of 
more intermediate steps), then again we put those steps in as hints.
With those additional hints,  then sometimes we could remove (one, some, or even all)
of the extra clauses
we added to the set of support and still get a proof.  If we did, we then 
replaced the hints that came from the proof obtained by using those additional clauses,
enabling that same proof to be found more easily.  

\paragraph{Divide and conquer.} 
If all the extra steps can be removed from the set of support, then we have the 
proof we were seeking.  But if not, then we wind up, for example, with a 
proof of the theorem from an extra hypothesis $P$.   At that point, we replace
the extra clause $P$ with its negation $-P$ in the set of support  and try 
again to find a proof, using the techniques described above, accumulating the steps
of any (partial) proofs found in the hints list.  Eventually (in practice) we 
always can succeed to find a proof from the extra hypothesis $-P$.  Now we have
 two proofs,  one from $P$ and one from $-P$.   If there had 
been two extra hypotheses $P$ and $Q$,  we might end up with four proofs, 
one from $P,Q$, one from $P,-Q$,  one from $-P,Q$, and one from $-P,-Q$.   
At that point we have ``reduced the problem to combining the cases.''

\section{Proof by cases}  \label{section:cases}
Proof by cases turned out to be difficult.  A number of proofs
in the book proceed by cases, for example according to whether $p$ lies on line $L$
or does not lie on line $L$.   Moreover, the lemma adjunction method 
often resulted in uncovering an intermediate step $A$ such that if we added
$A$ as an additional hypothesis, we could find a proof,  and we could also 
prove that step, by adding $-A$ as an additional hypothesis.  That amounts
to a proof by cases according as $A$ or $-A$.  It may seem that we must be 
nearly finished at that point.  But we found
that in such situations it was often not trivial, and sometimes
not possible,  to combine the two proofs into a single proof with no 
additional assumptions.  In this section, we discuss this difficulty.

 Let us suppose that 
we have a proof of the original goal from $A$  and another from $-A$.
What do we do?
We could of course just formulate two lemmas: $P$ implies the goal,
and $-P$ implies the goal. Then (in a third file), we could prove the goal from those two 
lemmas.  (Possibly just one of those lemmas would be enough.)
 Sometimes in a mathematics book, that is what is really done, even though 
the theorem is stated as a single theorem;  particularly when there is a certain 
symmetry to the theorem, instead of formulating a lemma explicitly, one simply says,
``without loss of generality we can assume.''  Still, we tried to eliminate the 
cases and give a proof in a single file, when we could do so.  We now 
describe our technique for that. 

\subsection{Tautology adjunction} 
   We could sometimes succeed by simply
adding a clause {\tt A | -A}, where the proof needs to proceed by cases on $A$.
\Otter seems to prefer constructive proofs!    This technique is 
called ``tautology adjunction'' by Wos, who used it 
decades ago in proving that subgroups of index 2 are normal.  We used this 
in many input files.  Here we discuss just one example. The inner connectivity
of betweenness (A17 above, Satz~5.3 in Szmielew) is derived as an 
easy corollary of Satz~5.1, which is
$$ a \neq b \land \T(a,b,c) \land \T(a,b,d) \implies \T(a,c,d) \lor \T(a,d,c).$$
 The natural 
way to formulate {\tt list(sos)} for this problem would be

\begin{verbatim}
     a != b.
     T(a,b,c).
     T(a,b,d).
     -T(a,c,d).
     -T(a,d,c).
\end{verbatim}

\noindent
Of course, that does not suffice to find a proof.  So, we added the 
description of the diagram, as given in \S\ref{section:diagrams}.  Unfortunately
\Otter could still not find a proof, even with hints.  

The proof in Szmielew proceeds by cases.   The methodology we followed 
in such cases was this:

\begin{itemize}
\item Add one case to sos, e.g. {\tt c=c1}.  ({\tt c1} is a constant from the diagram, above.)  
\item If we find a proof,  add the steps of that proof as hints.
\item Remove that case, and add its negation, e.g. {\tt c != c1}
\item If we find a proof,  add its steps also as hints.
\item Now remove both cases, and add their disjunction: {\tt c = c1 | c != c1}.
\end{itemize}

Often we could find a proof.   The example at hand required  two divisions into cases 
(so tautology disjunction was applied recursively).
 The first is the case whether {\tt c=c1} or not, 
and the second  whether {\tt d1=e} or not.  Here 
one can see in the two commented lines in {\tt list(sos)}  the 
traces of completing the last argument by cases. 

\begin{verbatim}
     % d1 = e.
     % d1!= e.
     d1=e | d1!=e.
     c = c1 | c != c1.
\end{verbatim}

We do not mean to imply that this was all there was to proving Satz~5.1.  This was just 
the last difficulty.  By that time, the input file already contained a long list of hints
obtained by methods described above.  The final proof had 127 steps.

\subsection{Remarks on tautology adjunction}
Suppose we have a resolution proof of contradiction from assumptions $L$ 
and the additional assumption $P$, and another proof 
from $L$ together with $-P$.  Then if we form the list of all clauses $Q | -P$,  where $Q$
is a clause in the proof from $P$,  we arrive at the conclusion $-P$, where we formerly
arrived at a contradiction.  Then we can continue with the proof of contradiction from $-P$.
This method, however, results in a proof only if the original proof from $P$ proceeded
by binary resolution.   If instead it used hyperresolution, then the resulting clauses 
$Q | -P$  do not form a proof, as the steps originally made by hyperresolution 
are no longer legal hyperresolution steps.  Steps by paramodulation might also be 
spoiled if the terms involved in the paramodulation occur in $P$.  These observations 
explain why it can be difficult to combine cases.  

At least we would like clauses of the form $Q | -P$ to be kept if they are deduced.
To achieve that when the $Q$ are in the hints list, we need (in \Ott)  to use {\tt hints}
instead of {\tt hints2}, and {\tt assign(fsub\_hint\_wt,-1)} to get clauses kept that are 
subsumed by hints.  Then we set {\tt max\_weight} very low, perhaps 6 or 8, which might be 
enough to allow the clauses $Q | -P$ to be deduced (in several steps) by binary resolution.
 Sometimes this approach worked, but {\tt hints} is a {\em lot} slower than 
{\tt hints2}.  (In Prover9,  {\tt hints} does not exist; what is called {\tt hints} is 
similar to {\tt hints2} in \Ott.)  Moreover, it did not always work (e.g., on Satz~12.1),
probably because some of the required inferences by binary resolution contain clauses that 
are not in the hints list, and so their weight is too large.  In principle there must be 
a setting of {\tt max\_weight} and a running time that produces the desired proof, but 
it sometimes eluded us.

In principle it should be possible to implement an algorithm for combining two proofs
(one from $P$ and one from $-P$)  and producing a valid proof by binary resolution and paramodulation.
Then the steps of that proof could be injected as hints.  The resulting proof 
would not rely for its correctness on the algorithm for combining proofs,  since if the 
hints work, it doesn't matter where they came from.   But we have not 
tried to implement such an algorithm.  We note, however, that such an algorithm could 
be implemented ``on top of'' existing theorem provers, without modifying the theorem prover,
using PHP scripts (or scripts written in some other language). 

\section{The challenge problems} \label{section:results}

All the input files and resulting \Otter proofs that we found are posted 
on the web at \cite{tarski-archive}.     Here
we list some of the more difficult proofs we found, as well as some of interest
for reasons other than difficulty.

\subsection{Properties of betweenness}
In \S\ref{section:betweenness}, we listed six difficult theorems (A12-A18 except A13),
each of which Tarski originally took as an axiom.  They must have been fairly
difficult if Tarski did not notice that they could be eliminated as axioms.
(A13 disappears when equality axioms are included; in 1926 equality axioms were not
included.)
All six occur  as theorems
in \cite{schwabhauser}. 
We found \Otter proofs of all those theorems, several of them  short in spite
of the perceived difficulty of the theorems. 
Table~\ref{table:1} gives the
length of the proofs we found, which is perhaps some indication of the relative 
difficulty of finding them. We found the short proofs completely mechanically.
Satz~5.1, with 121 steps, is a theorem from Gupta's Ph.D. thesis \cite{gupta1965},
and it is the first theorem in Szmielew's development that we could not prove 
completely mechanically.  We had to use lemma adjunction and hint injection.

\begin{table}
\caption{Proofs found for some of Tarski's original axioms}
\label{table:1}
\center{
\begin{tabular}{l l l r l}
A12 \ \ & Satz 3.1 \ \ & Reflexivity for betweenness & 4 steps \\   
A14 & Satz 3.2 & Symmetry for betweenness &  4 steps \\
A15 & Satz 3.5 &  Inner transitivity & 4 steps\\
A16 & Satz 3.7 & Outer transitivity & 16 steps\\
A17 & Satz 5.3 & Inner connectivity & 15 steps\\
A18 & Satz 5.1 & Outer connectivity & 121 steps
\end{tabular} }
\vskip -0.3cm 
\end{table}

Another theorem about betweenness   occurs as a 
challenge problem in \cite{wos1988}, namely, the ``five-point theorem'':
$$ \T(z,w,v) \land \T(z,y,v) \land \T(w,x,y) \implies \T(z,x,v).$$
This  theorem does not occur in Szmielew's manuscript, 
but it has a proof with 6 steps using the theorems mentioned above,
found completely mechanically.

\subsection{Midpoints, perpendiculars, and isosceles triangles}
The ``midpoint theorem'' asserts that every segment has a midpoint.  The traditional Euclidean 
construction involves the intersection points of two circles, but we are required to 
prove the theorem from A1-A9.  (Not even the parallel axiom A10 is to be used).  
This is a difficult problem, and was apparently not solved until Gupta's 1965 thesis
\cite{gupta1965}.  Two  important preliminary steps are the erection of a perpendicular
to a line at a given point, and the ``Lotsatz'', which says we can drop a perpendicular
to a line from a point not on the line.  Remember this must be done without circles!
 A clever observation of Gupta was 
that one can fairly easily construct the midpoint of $ab$ if $ab$ is the base of an isosceles
triangle (only two applications of inner Pasch are needed).  This plays a key role in 
the proof of the Lotsatz.   The two theorems on perpendiculars are used to construct
the midpoint.  Of course, once we have midpoints and perpendiculars, it is trivial to 
show that every segment is the base of an isosceles triangle; that theorem does not 
even occur explicitly in \cite{schwabhauser}.  An important lemma used in the proofs
of these theorems is the  ``Krippenlemma'', a key result in Gupta's thesis.  For a diagram 
and formal statement, see Lemma~7.22 of \cite{schwabhauser}, pp.~53--54.  
Table~\ref{table:2} shows the lengths of our proofs of these theorems, obtained using lemma adjunction and hint injection.
 
\begin{table}[ht]
\caption{Proofs of Gupta's theorems found using lemma adjunction and hint injection}
\label{table:2}
\center{
\begin{tabular}{l l r l}
 Satz 7.22 \ \ & Krippenlemma & 96 steps\\
 Satz 7.25 & Base of isosceles triangle has a midpoint &  113 steps\\
 Satz 8.18 & Lotsatz: there is a perpendicular to a & \\
 & line from a point not on the line & 227 steps \\
 Satz 8.21a & There is a perpendicular to a line through  \\
 &  a point on the line on the opposite side from & \\
 &  a given point not on the line. & 106 steps \\
Satz 8.24b & Given segment $ab$ and perpendiculars $ap$ and $qb$,    \\
& and point $t$ on line $ab$ between $p$ and $q$, \\
&with $ap\le qb$, then  segment $ab$ has a  midpoint. & 201 steps \\
Satz 8.22 & Every segment has a midpoint & 22 steps
\end{tabular} }
\vskip-0.3cm
\end{table}
\medskip

We obtained a 108-step proof of the Krippenlemma using the subformula strategy, which 
works without reference to the steps of the book proof.  It took 3,189 seconds to find.
We were surprised to find such a long proof completely mechanically!   
None of the other theorems listed here could be 
proved by the subformula strategy in one hour, although, 
as described in \S\ref{section:hardtheorems}, we did find several more proofs of 
other difficult theorems.

\subsection{The diagonals of a rhomboid bisect each other}
A rhomboid is a quadrilateral whose opposite sides are equal.
One of the challenges in \cite{wos1988} (see p.~214) solved by Quaife,
was to prove that the diagonals of a rectangle bisect each other. 
A more general problem is found in Satz~7.21, which asserts that 
if the diagonals of a rhomboid meet, then they bisect each other.  Quaife
also proved this theorem with \Ott. 
Our proof of this theorem has 26 steps.  
Note that no upper-dimension axiom is used, so it is necessary to assume the diagonals 
meet, since otherwise the theorem fails in 3-space.  Even though 26 steps is fairly short,
we could not find this proof completely mechanically, even with the subformula strategy. 

\subsection{Inner and outer Pasch}
The proof that inner Pasch implies outer Pasch (using A1-A6 and A8) was one of the 
major results of Gupta's thesis, and enabled Szmielew to replace outer Pasch by 
inner Pasch as an axiom.  This theorem was one of Quaife's four challenges.  It
is Satz~9.6 in \cite{schwabhauser}.   The proof posted on our archive is 111 steps,
preceded by proofs Satz~9.4 and Satz~9.5 of 57 and 45 steps, respectively. Satz 9.5 is the ``plane
separation theorem'',  important in its own right.  

\subsection{Hilbert's axioms}  
Hilbert's theory can be interpreted in Tarski's, using pairs of 
points for lines and ordered triples of points for angles and planes.   His 
axioms (so interpreted)
 all turn out to be either axioms, theorems proved in \cite{schwabhauser}, or 
extremely elementary consequences of theorems proved in \cite{schwabhauser}.   
  The theorems of \cite{schwabhauser}
needed are 2.3,2.4,2.5,2.8; 3.2,3.13;6.16, 6.18; 8.21, 8.22; 9.8, 9.25, 9.26, 11.15, 11.49;
and   Hilbert's parallel axiom is Satz 12.11.  We have posted \Otter proofs of all these
theorems.  Narboux and Braun have proof-checked Hilbert's axioms in Tarskian geometry, using Coq \cite{narboux2012}.

\section{1992 {\em vs} 2015: improved strategies or faster computers?}
  
The research reported here shows how much progress has occurred in automated reasoning in that 
time period.  Indeed, approximately thirty years ago, almost all of the theorems cited in this article were out of reach.
The question arises whether this advance might be due simply to the increased memory capacity and 
speed of modern computers.   Perhaps Quaife, equipped with one of our computers, would have found 
these proofs?  Perhaps we, constrained to run on a computer from 1990, might not have found them?
We argue that this is not the case:  the improvements are due not to faster hardware but to 
the techniques described above, namely generating partial proofs (of intermediate steps) and using 
their steps as hints; using the right combination of inference rules and settings; using 
tautology adjunction to help with proofs by cases; divide-and-conquer; and finally,
the spectacular success of the subformula strategy.    We note that Quaife
did have (in fact, invented) the technique of giving \Otter the diagram.  We did not actually try 
to run on a 1990 computer, and we do not doubt that it would have been painful and discouraging;
but we think the main credit should go to Veroff's invention of hints, and the uses of hints
developed by Wos and applied here.  Proofs of length more 
than 40 were out of reach in 2014 for completely mechanical derivations, and could then only 
be found by lemma adjunction and hint injection.  But in 2015 several proofs of length more 
than 90 were found completely mechanically by the subformula strategy,  using the same 
computer we used in 2014.

\section{Proof checking {\em vs.} proof finding}
``Proof checking'' refers to obtaining computer-verified proofs, starting with human-written proofs.
``Proof finding'' refers to the traditional task of automated deduction, finding a proof by searching a large space of possible proofs,  either without possessing a proof or without making use of a known proof.
Our work shows that this distinction is not as clearcut as it might seem. 
  If we have a proof in hand (whether generated by 
human or machine), and we enter its steps as hints, with a low {\tt max\_weight}, we 
force \Otter to find a proof containing mostly the same formulas as the proof in the hints.  
(The order of deductions might be different.)   This happens immediately if the steps
were machine steps, and we have explained above the techniques we use if the original steps
are human proof steps.   By those methods, we can almost always ensure that \Otter finds 
a proof, if we have a proof in hand.  One could plausibly claim that this is proof checking,
not proof finding. 

On the other hand, today's proof checkers are increasingly incorporating (calls to) external
theorem provers to help reduce the number of steps a human must supply.   If a theorem prover
can get from the existing proof to the next goal,  then the returned proof can often 
be translated into the proof checker's language and checked.  For example, the Sledgehammer
program \cite{paulson2010} is used with Isabelle, and HOLY-Hammer \cite{holyhammer}
 is used with HOL Light. 
In this way, proof checking is incorporating elements of automated theorem proving. 
See \cite{blanchette2015} for a survey of this research and further references.

\section{Developing an entire theory using a theorem prover}
When we made our first report on this work \cite{beeson2014-wos}, we 
had two hundred hand-crafted input files,  one for each theorem.   These files 
had been made by cutting and pasting lists of axioms,  adding the positive form 
(Skolemization) of the last theorem by hand,  and often had been heavily edited
during the process of applying our proof-finding methods.  There was plenty
of room for human error in the preparation of these files; and as it turned out 
in 2015,  they contained numerous errors;  none irreparable, but one wishes 
for a standard of perfection in computer-generated proofs.  

Thinking about the sources of error and the means to remedy them,  we realized
that a major source of error could be the hand-translation from the negated
form of a theorem (used in the set of support to prove it) to the ``positive form''
or Skolemization,  in which the theorem is entered into subsequent input files 
for use in proving further theorems.   We therefore prepared a ``master list''
containing, for each theorem,  the positive and negative forms and, in addition,
the ``diagram equations'' defining the diagram for that theorem, if any.
The master list also contains the names of Skolem functions to be used in 
Skolemizing that theorem,  if the theorem contains (existential) variables.
We also optionally included a list of tautologies that were needed to prove 
that theorem,  if we could not eliminate the tautologies.

The plan was that the master list could then be used to mechanically generate 
input files.   The correctness issues would then have been conveniently divided:

\begin{itemize}
\item  Is the positive form of each theorem in the master list really the correct
Skolemization of the negative form?
\item Does the diagram entry in the master list have the form of an equation with a new 
constant on the left?  And on the right, does it mention only previously-introduced Skolem functions?
\item Is each Skolem function introduced in exactly one theorem? 
\item  Does the master list actually correspond to the statements of the theorems in the book?
\item  Are the input files being correctly generated?
\item  Do they all produce proofs?
\item  And of course:  are the proofs produced by \Otter correct? (This is not a novel
issue and will not be discussed.)
\end{itemize}

The question whether the diagram in the master list corresponds to the book's
diagram is not relevant.  If one gets a proof with any diagram whatsoever (meeting the 
above conditions),  it is fine.
Adding equations defining a new constant is conservative; that is,  no new theorems can be 
proved by adding such axioms.  However, the meaning of ``new constant'' is not quite
as obvious as one might think.  Consider, for example, the diagram for Satz 3.17.
There are two equations:
$$ e = ip(c,b1,a1,a,p)$$
$$ d = ip(c,b,a,b1,e)$$
The new constants are $d$ and $e$.  The equation for $d$ has $e$ on the right, but 
that is okay.  What would be wrong is if the equation for $d$ had $d$ on the right,
or if the equation for $e$ had $d$ on the right and the equation for $d$ had $e$ on the right.
What we mean by ``new constant'' is  ``constant not occurring in the theorem or the 
right sides of the previous diagram equations.''

Similarly, it does not matter what one puts in the hints list.  If one gets a proof,
it is a proof, regardless of what was in the hints list.  In our mechanical generation 
of input files,  we make use of the (possibly unreliable) proofs we found in 2012.
If we cannot find a proof without hints, then we use the steps of an existing (alleged)
proof as hints.   Thus, the sources for generating input files are 

\begin{itemize}
\item The master list
\item The existing (alleged) proofs
\item The program that produces the input files from the master list and proofs
\end{itemize}

\subsection{The master list}
The master list is a text file, a piece of code in the PHP programming language.
That file begins with class definitions of classes called {\em Theorem}, {\em Definition},
and {\em Axiom}.  The class {\em Theorem}, for example, looks like this (but 
for simplicity we here omit the constructor function).  (Those unfamiliar with PHP
should just ignore {\em var} and the dollar signs.)

\begin{verbatim}
class Theorem
{ var $name;
  var $PositiveForm;     // array of clauses 
  var $NegatedForm;      // array of clauses
  var $Diagram;          // array of diagram equations
  var $SkolemSymbols = ""; // array of Skolem symbols introduced
  var $Cases = ""; // cases we used to split the problem, if any.
}
\end{verbatim}

\hyphenation{Tarski-Theorems}
\noindent
After that come arrays defining {\em TarskiAxioms}, 
{\em TarskiDefinitions}, and  {\em TarskiTheorems}.  
The {\em TarskiTheorems} array is the heart of the master list.
It starts off like this:

\begin{verbatim}
$TarskiTheorems = array(
	new Theorem( "Satz2.1", 
		array("E(xa,xb,xa,xb)"), 
		array("-E(a,b,a,b)")
	),  
	new Theorem( "Satz2.2", 
		array("-E(xa,xb,xc,xd) | E(xc,xd,xa,xb)"),
		array("E(a,b,c,d)","-E(c,d,a,b)")
	), 
\end{verbatim}

\noindent
As one can see, the master list is human readable, not much encumbered
by the occurrences of {\tt new} and {\tt array} that make it into PHP code.
The master list was prepared by hand, by the following method:
First, we entered the negated form of a theorem, by copying 
it from {\tt list(sos)} of the existing hand-crafted input file.
It was, we realized, important that the constants in the master list
have the same names as the constants in the corresponding 
existing proofs,  to avoid incompatibilities when those proofs are
used as hints. (Of course, that  matters only for the theorems that
cannot be proved without hints.)%
\footnote{Therefore we could not satisfy a request that the constants
have names unique across all files.  Researchers with that wish can 
write PHP code to append a number to each constant.  Then, however, they 
won't be able to use hints extracted from our proofs.
}

Then, we computed (by hand) the positive form. 
Here constants such as $a$ were changed to variables such as $xa$.
In some cases we had constants $cu$ or $cx$, where the book proof 
had $u$ or $x$;  in computing the positive form these become $u$
or $x$ again, rather than $xcx$ or $xcu$.  Although this was
originally done by hand, we checked it (and other things about the 
master list) with a PHP program {\em TestMasterList.php}.  This program
carries out Skolemization  on a text-only basis, without parsing
terms or clauses.  (Any parse errors will turn up when we run the resulting input files.)
The program copes correctly when the negative form has one clause that is a disjunction.
In the entire Tarski corpus, there are only two theorems whose negative form contains
two disjunctions.  These were Skolemized by hand and checked carefully.
That the resulting input files do result in proofs is another positive indication.
This mechanical check of correct Skolemization is our answer to the first
itemized correctness concern above.

One issue was the order of arguments to Skolem functions.  In 2015 we could 
not change the conventions that we adopted in 2013, since that would have rendered the 
hints obtained from our existing proofs useless.  That necessitated some 
not very beautiful code that specifies the order per Skolem function.  The correctness
of this code is evidenced by the fact that it did work with hints obtained from 
existing proofs.  Luckily, in 2013 when we worked on one file at a time, we were consistent
about the order of arguments of the same Skolem function in different files.

The second concern was whether each diagram clause has the form of an 
equation with a new constant on the left, in the precise sense defined above.
Our program {\tt TestMasterList.php} checks this condition mechanically.  It 
also checks that each Skolem function is introduced by exactly one theorem and 
is not used (even in a diagram) before it is introduced.

The third concern was whether each theorem in the master list corresponds
to the version in the printed book.  Several considerations arise here:
First, Szmielew treated lines as sets of points, so her treatment is not 
strictly first order. We translated ``line $L$'' to two points $p,q$ and 
the literal $p \neq q$, and $x \in L$  to $Col(p,q,x)$,  where $Col$ means
``collinear'' and is defined in terms of betweenness.  The correctness issue
has to be understood modulo this translation.  Second, as a result of that
treatment of lines,  some simple theorems had to be formulated in our development
that are not in the book.  For example, perpendicularity of lines becomes
a 4-ary relation $perp(a,b,c,d)$, or $ab \perp cd$,  and we have to prove that, for example, 
$a$ and $b$ can be interchanged and $c$ and $d$ can be interchanged, or both.
This gives rise to some theorems that do not occur in the book.  But for those
theorems that do occur in the book, one can check by hand that they do correspond 
to the book.
 Since the Skolemization has been checked by machine,
it is only necessary to check that the negative form corresponds to the book.
There are only a few cases where this is at all problematic.  Lemma~9.4, p.~68
is an example.  The question whether the formal theorem corresponds to 
the intended theorem arises also in using a proof checker. One
cannot check by means of a computer program that the master list
 corresponds to the book.  Mathematics books
written in the twentieth century do not contain machine-readable formulas,
and even if they did,  we would still need to know if those really represented
Szmielew's intent.  (There are a few typographical errors in Szmielew.)

\subsection{Size and availability of the master list}
We briefly contemplated including the master list as an appendix to this paper.
We discovered that, when printed, it is 43 pages long.  Moreover, in order
to be useful, it must be in electronic form.  It could well turn out that 
the master list is the most useful product of this research, since it can be 
used immediately by anyone else wishing to study the Tarski axioms and Szmielew's
axiomatic development from them.  We therefore plan to post a separate
document to the ArXiv containing the master list in the \TeX\ source.  Then,
as long as the ArXiv lives, the master list will be available. Of course
the master list, as well as the PHP programs used for this research, will be
available at \cite{tarski-archive} for at least the next few years.

\subsection{Generating input files mechanically} \label{section:mechanicalinputfiles}
Once the master list is correct, it is a routine programming exercise
involving only textual (string) manipulations to generate input files.
The algorithm is simple.  Consider, for example, the function
with the specification given in the following comments.

\begin{verbatim}
function InputFileAsArray($t, $diagrams, $sos, $settings, $cases)
// $t should be a member of $TarskiTheorems
// Generate the lines of an input file for proving $t
// and return an array of strings containing those lines.
// Make sure no line exceeds 80 characters.
// include the diagram if $diagrams is not false.
// include the members of $t->Cases if it is an array
\end{verbatim}

Here {\tt \$settings} is an array of the \Otter settings to be used
in this file; the strings in the arrays
 {\tt \$t->Diagrams}, {\tt \$t->NegatedForm }, and {\tt \$t->Cases}
are ready to be copied out of {\tt \$t} into the appropriate places in a file
template.   All the function has to do is 
create an array {\tt \$ans} and add these things to that array in the right
order,  interspersing template lines like ``{\tt list(sos).}''  at the appropriate places.

This function, however, doesn't put any hints into the file it is creating.
That is done by the following code.

\begin{verbatim}
function InsertHintsIntoInputFile($theorem_name, $outdir, $indir)
// Get hints from the proof, i.e. from the .prf file  
// found in directory $outdir, and insert them in 
// list(hints) in the input file.
\end{verbatim}

The heart of this program is {\tt GetHintsFromProof},  which we have been 
using since 2012 to extract hints by hand from a proof; we have confidence 
in the correctness of that program based on long use and on many comparisons
of the output with the input.  Again, the correctness of that program doesn't
really matter:  whatever we put into the hints list, if we get a proof,
the proof is right.   In order to ensure correctness of any proofs obtained, 
the only thing that really matters about {\tt InsertHintsIntoInputFile}
is that all it does to the input file is add a hints list.  (Of course,
to actually obtain some proofs, it had better be correct in other ways.)

\subsection{Experimenting with many files at a time}

With the aid of the PHP programs discussed above, we wrote short
PHP programs that test a given set of files (such as, for example, those belonging to Chapter 8),
with a given set of settings for the theorem prover, and given values of variables
controlling whether we will use hints extracted from a proof or not, whether we will 
use diagrams, whether we will use the case listed in the master list,
whether we will try to prove theorems that have already been proved or only ones
that have not been proved yet, and so forth.  We also could have the settings used on each 
input file depend on previously-obtained results, the record of which was kept in 
a machine-readable (but handwritten) file.   The PHP tools that we used are
 available, with descriptions, at \cite{tarski-archive}. 

\subsection{Summary of the correctness discussion}
We were able to replace the hand-crafted input files we posted in 2013--2014
by mechanically generated input files.  In the case of difficult theorems that 
we did not prove completely mechanically, our mechanically generated files used
 hint injection,  with hints extracted
from the old proofs.   Sometimes, if the proof was shorter, we iterated,  injecting
hints obtained from the new proof.   In most cases the resulting proofs were
shorter than the ones we had before.  As discussed above, if our PHP programs for 
generating those files are correct and the Skolemizations in the master list are correct,
 then each theorem has been deduced from the 
ones preceding it in the master list.  In other words, it is a conclusion, not a 
hypothesis, that the theorems are ordered correctly in the master list.   
The correctness of the Skolemizations in the master list has been machine-checked
(except for the one theorem with two disjunctions, which was checked manually).  
That the master list correctly corresponds 
to the book has been checked manually (the only possible way to check it).

\section{Easy and hard theorems and their proofs} \label{section:hardtheorems}
We divide the Tarski theorems into ``easy'' and ``hard'' theorems, according to the somewhat
arbitrary rule that a theorem is ``easy'' if it has a proof of 40 or fewer steps. The idea
behind the choice of the number 40 is
 that we were able to prove many of the easy theorems completely automatically,
while to prove the hard theorems,  we had to use hints.   There are 214 
theorems all together, of which 29 are ``hard'' by this definition.  We proved about 
two-thirds of the 
easy theorems mechanically, and we proved four very hard theorems completely mechanically
too, using the subformula strategy,  leaving 25 theorems that required hints.  
Here ``mechanical'' refers to proofs
 obtained with no 
reference to the book proof. An additional 14 proofs were found by ``giving the prover
the diagram'',  without additional use of the book proof.   That raises ``about two-thirds''
to ``about three-quarters''.
 
The rest of the easy proofs, and all the hard proofs, 
were proved using ``hints''.  These hints came from using some steps from the 
book proofs, and from  proofs of related theorems obtained by the 
methods described above (lemma adjunction and eliminating cases).  

If we had to measure the strength of \Ott\ on this problem set by a single number, we would say 
that 40 is the answer:  we expect that \Ott\  has a good chance to automatically find 
proofs for theorems that have proofs of length less than or equal to 40,  and
 only occasionally to find 
proofs for theorems much longer than 40.

The book starts   with easy theorems.  Chapters 2 through 6 contain 81 
theorems.  Of those only two are hard:  Satz 4.2 (44 steps) and Satz 5.1 (127 steps).
  Satz~5.1 expresses in Tarski's language that 
if we fix $a$ and define $x \le y$ by $\T(a,x,y)$, then $x \le y \lor y \le x$.  
This was first proved from Tarski's axioms in Gupta's thesis \cite{gupta1965}.
We had to use diagrams on six of the remaining theorems.  We 
 used hints on two ``easy'' theorems:   Satz~4.16 (23 steps) 
and Satz~5.12b (16 steps).  We could prove Satz~4.16 without hints,
but we had to wait 95696.25 seconds (more than 26 hours).
Satz 4.5 required the subformula strategy in order to find a proof automatically.
Satz 6.16b appeared to be hard at first, but the 
subformula strategy found a short proof.   Of the remaining 69 theorems, we proved 67 
completely mechanically, without hints or diagrams, using our default settings.   The 
other two are Satz 3.1 and Satz 4.6.  The former we could prove mechanically
with a diagram,  but we could also prove it without a diagram if we put everything
in the set of support, instead of putting the axioms in list(usable).  Satz~4.6,
strangely enough,  could be proved immediately after adding two instances of 
equality axioms explicitly to the set of support,  but we could not get a proof 
without adding those equality axioms.  Of course, once we found the proof by adding
equality axioms, we could put the steps of the proof in as hints and find a proof
without explicit equality axioms. 

In Chapters 7 and 8, there are 46 more theorems that we proved completely mechanically,
8 more hard theorems, 3 short theorems on which we had to use hints (i.e., could 
not prove mechanically even with a diagram),  and five theorems that we could 
prove mechanically with a diagram, but not without.   Of those eight hard theorems,
two were proved mechanically using the subformula strategy: Satz 7.13 and 
Satz 7.22a.  The latter of these is the Krippenlemma,  one of the important theorems
of Gupta's thesis.  That a 108-step proof of a major result could be found 
completely automatically was completely unexpected. 

Chapter 9 gets harder.  We could prove only four theorems  completely mechanically;
diagrams were not needed and did not help with the other theorems.  There are five hard theorems,
which we proved using hints. 
We were also able to prove two of those hard theorems with the subformula strategy,
including the gem of Gupta's thesis, that outer Pasch implies inner Pasch.  
There are also five short theorems that we could prove only by using hints, although
four of them have proofs between 35 and 40 steps, so they are ``borderline short.''

Chapters 10 gets easier again;  it is about reflection in a line
and right angles, and the high point is the proof of Euclid's fourth postulate, that 
all right angles are congruent, in the form that two right triangles with congruent legs
also have congruent hypotenuses.  Although five proofs in Chapter 10 are longer
than 45 steps, only two are longer than 75 steps.  The chapter ends with a proof 
of Hilbert's triangle construction axiom.

Chapter 11 deals with the definition and properties
of angles (as triples of points).  Only two proofs in Chapter 11 are longer than 40 steps.

Table~\ref{table:3} summarizes the results just described.%
\footnote{There are 214 theorems in our Master List; but the last three are all the 
Hilbert parallel axiom, in two cases and in a combined statement. So really, there are 212
theorems, of which 29 are hard and 183 are easy.}

\begin{table}
\caption{One hundred eighty three theorems with proofs of length $\le 40$}
\label{table:3}
\center{
\begin{tabular} { l r r r r}
{\bf Chapter} & {\bf no diagram} & {\bf diagram} & {\bf Subformula strategy} & {\bf  hints}   \\
2 to 6 & 72 & 5 & 3 & 2\\
7 & 20 & 0 & 1 & 1  \\
8 & 16 & 6 & 2 & 2  \\
9 & 5 & 0 & 3 & 2  \\
8A & 6 & 1 & 0 & 1  \\
9A & 6 & 0 & 0 & 4  \\
10 & 6 & 1 & 0 & 3  \\
11 & 3 & 1 & 0 & 6  \\
12 & 0 & 0 & 0 & 5  \\
Total & 134 & 14 & 9 & 26
\end{tabular}
}
\end{table}

\begin{table}
\caption{Twenty-nine hard theorems and their proofs}
\label{table:4}
\center{
 
\begin{tabular} {l l r r r }
  {\bf Theorem} & {\bf description} & {\bf sub. str.} & {\bf hints}  & {\bf book }\\
  Satz 4.2  &  inner 5-segment theorem && 44  & 13\\   
 Satz 5.1 & trichotomy &&  127 &  70 \\
  Satz 7.13 & refl. in pt. preserves congruence  & $(99, 921)$ & 72  & 28\\
 Satz 7.22a & Krippenlemma & $(108,3189)$& 96 & 27 \\
 Satz 7.25 & isosceles triangle has midpoint && 113 &34\\
 Satz 8.18 & Lotsatz (dropped perpendicular)  && 227 & 32\\
 Satz 8.21a & uniqueness of dropped perp   && 106& 5\\
  Satz 8.22b & existence of midpoint   &&201 & 42\\
   Satz 8.24a &  midpoint   details      && 171& 42\\
    Satz 8.24b &  midpoint details     && 163& 42\\
  Satz 9.3 & opposite-side ray theorem   && 52 & 14\\
  Satz 9.4b & $r=s$ case of 9.4       &&44 &7\\
    Satz 9.4c & second half of Lemma 9.4 in book       &&41 & ?\\
   Satz 9.6 & inner Pasch implies outer Pasch  &$(98,186)$ &91& 27 \\
    Satz 9.8 & opposite, same implies opposite  &$(63,1646)$ &48 &21\\
    Satz 9.13 & transitivity of samesideline   && 71& 1?\\
    Lemma 9.13f & (case 2)                     && 42 & ? \\
     Satz 10.2a & existence of reflection in line  && 45 & 23\\
     Satz 10.2b & uniqueness of reflection in line  && 49& 23\\
     Satz 10.10 &  refl. in line preserves congruence                    && 72 & 26\\
     Satz 10.12a & 	right triangles with common vertex & & \\
     & $\ldots$ have congruent hypotenuses & & 60 & 6\\
      Satz 10.12b &  right triangle theorem (Euclid 4)  && 91 &14\\
      Sata 10.15 &  perpendicular on given side of line && 74 & ? \\
      Satz 10.16a & triangle construction  && 57 & 24?\\
      Satz 10.16b & triangle uniqueness   && 60&22\\
      Satz 11.3a & angle congruence theorem  && 43& 18\\
      Satz 11.15a &   angle transport(existence)  && 78 & 1? \\
      Satz 11.15b &   angle transport (uniqueness)  && 78 & 1? \\
      Satz 12.11  &  Hilbert's parallel axiom && 105 & 18?
\end{tabular} }
\end{table}

\FloatBarrier

Table~\ref{table:4} shows the hard theorems from the entire book,  with the 
number of steps in the proof(s) that we obtained.  A paired number like $(99,921)$ 
indicates  
a proof found mechanically using the subformula strategy, of length 99 and found in 921 seconds.
The last column is the number of steps in the book proof.  A question mark in this column indicates
that the book proof contains non-trivial gaps or is omitted. 
Not counting the lines with a question mark, we have 2032 \Otter steps
corresponding to 516 book steps, giving a de Bruijn factor (the ratio of the two) 
of 4. 

Of the 212 theorems we tried to prove, we proved 147  completely automatically 
(without any use of the book proof or diagram);   we needed the 
subformula strategy for 13 of those proofs.   We were able to prove 15 more theorems
by using the diagram, without using the steps of the book proof.  And when we allowed
ourselves to use some of the steps of the book proof as hints,  we achieved a 100\% success rate.

\section{What about Prover9? Or E, {\textsc SPASS}, Vampire?}

Nearly every person with whom 
we have discussed this work, including the referees, has suggested that another prover
or provers might do better in some respect than \Ott.  It was not our purpose 
to evaluate the performance of any theorem prover or to compare the performances of 
different theorem provers.  Our purpose was to show that a combination of strategies 
in {\em some} theorem prover could prove all the approximately 200 theorems in Tarski
geometry.

 In \cite{narboux2015b},  Durdevic {\em et.al.} tried E, {\textsc SPASS}, and Vampire
on the theorems from Szmielew, with a 37\% success rate.   The corresponding 
percentage in our work, as discussed in the previous section, was 62\% without the subformula strategy, and 
76\% with the subformula strategy.
 These numbers may not be exactly comparable, since the set of theorems may not have been identical.
They used a very low time
limit,  but most of the theorems we proved mechanically took less than 20 seconds, so 
the few that took longer did not change our percentage much.   We attribute our higher 
success rate (without any strategy at all)  
to a good choice of settings for \Ott.   The fact that we did achieve a 100\% success
rate with the use of lemma adjunction and hint injection  is another reason for not using
another prover:  we could not improve on that success rate.

Our master list of the Tarski theorems is now available for others to use; also our 
PHP programs for manipulating the master list and files are available.  Conducting experiments
with other theorem provers then involves only modifying the template for an input file
that is used by {\em InputFileAsArray}.  
  We have not conducted such experiments, but others are doing so as this article goes
to press,  and will report on their experiments in due course.  We note that in order
to duplicate our result with \Ott (134 out of 212 proved mechanically),
 it was necessary to use the default settings we used 
for \Ott, which can be found in the PHP code supplied on our website.   
Tentative results for Vampire and E are 154 and 142 out of 212.

\section{Summary}
\vskip-0.3cm
We used \Otter to find proofs of the theorems in Tarskian geometry in the first nine 
chapters of Szmielew's development in Part I of \cite{schwabhauser}.  Those theorems 
include the four unsolved challenge problems from Quaife's book\cite{quaife1992}, and 
the verification of Hilbert's axioms.  Of the 214 theorems, we proved
62\% mechanically and without strategies (except for a good choice of settings), 
and by the use of the subformula strategy we increased that to 76\%.  By using 
lemma adjunction and hint injection, which depend on starting with a book proof, 
we achieved a 100\% success rate.

Mechanically generated input files and the resulting proofs for all the theorems that we have proved  
are archived at \cite{tarski-archive}.



%
 
%

\begin{acknowledgements}
This material was based in part on work supported by the U.S. Department of Energy,
Office of Science, under contract DE-ACO2-06CH11357.
\end{acknowledgements}


\begin{thebibliography}{10}
\providecommand{\url}[1]{{#1}}
\providecommand{\urlprefix}{URL }
\expandafter\ifx\csname urlstyle\endcsname\relax
  \providecommand{\doi}[1]{DOI~\discretionary{}{}{}#1}\else
  \providecommand{\doi}{DOI~\discretionary{}{}{}\begingroup
  \urlstyle{rm}\Url}\fi

\bibitem{tarski-archive}
Beeson, M.: The {T}arski formalization project\hfill.
\newblock {h}ttp://www.michaelbeeson.com/research/FormalTarski/index.php

\bibitem{beeson2014-wos}
Beeson, M., Wos, L.: {OTTER} proofs in {T}arskian geometry.
\newblock In: S.~Demri, D.~Kapur, C.~Weidenbach (eds.) 7th International Joint
  Conference, IJCAR 2014, Held as Part of the Vienna Summer of Logic, Vienna,
  Austria, July 19-22, 2014, Proceedings, \emph{Lecture Notes in Computer
  Science}, vol. 8562, pp. 495--510. Springer (2014)

\bibitem{blanchette2015}
Blanchette, J.C., Kaliszyk, C., Paulson, L.C., Urban, J.: Hammering towards
  {QED}.
\newblock Journal of Formalized Reasoning  (2015, to appear)

\bibitem{narboux2012}
Braun, G., Narboux, J.: From {T}arski to {H}ilbert.
\newblock In: T.~Ida, J.~Fleuriot (eds.) Automated Deduction in Geometry 2012,
  pp. 89--109 (2012)

\bibitem{narboux2015}
Braun, G., Narboux, J.: A synthetic proof of {P}appus's theorem in {T}arski's
  geometry.
\newblock Journal of Automated Reasoning  (2015(?) to appear)

\bibitem{cavinessjohnson}
Caviness, B.F., Johnson, J.R. (eds.): Quantifier Elimination and Cylindrical
  Algebraic Decomposition.
\newblock Springer, Wien/New York (1998)

\bibitem{narboux2015b}
Durdevic, S.S., Narboux, J., Janicic, P.: {Automated Generation of Machine
  Verifiable and Readable Proofs: A Case Study of Tarski's Geometry}.
\newblock {Annals of Mathematics and Artificial Intelligence} p.~25 (2015).
\newblock \doi{10.1007/s10472-014-9443-5}.
\newblock \urlprefix\url{https://hal.inria.fr/hal-01091011}

\bibitem{gupta1965}
Gupta, H.N.: Contributions to the axiomatic foundations of geometry.
\newblock Ph.D. thesis, University of California, Berkeley (1965)

\bibitem{hilbert1899}
Hilbert, D.: Foundations of Geometry ({G}rundlagen der {G}eometrie).
\newblock Open Court, La Salle, Illinois (1960).
\newblock Second English edition, translated from the tenth German edition by
  Leo Unger. Original publication date, 1899.

\bibitem{holyhammer}
Kaliszyk, C., Urban, J.: Learning-assisted automated reasoning with {F}lyspeck.
\newblock Journal of Automated Reasoning \textbf{53}(2), 173--213 (2014).
\newblock \doi{10.1007/s10817-014-9303-3}.
\newblock \urlprefix\url{http://dx.doi.org/10.1007/s10817-014-9303-3}

\bibitem{pasch1882}
Pasch, M.: {V}orlesung \"uber {N}euere {G}eometrie.
\newblock Teubner, Leipzig (1882)

\bibitem{pasch1926}
Pasch, M., Dehn, M.: {V}orlesung \"uber {N}euere {G}eometrie.
\newblock B.~G. Teubner, Leipzig (1926).
\newblock The first edition (1882), which is the one digitized by Google
  Scholar, does not contain the appendix by Dehn.

\bibitem{paulson2010}
Paulson, L.C., Blanchette, J.C.: Three years of experience with {S}ledgehammer,
  a practical link between automatic and interactive theorem provers.
\newblock In: G.~Sutcliffe, S.~Schulz, E.~Ternovska (eds.) IWIL 2010. The 8th
  International Workshop on the Implementation of Logics, \emph{EasyChair
  Proceedings in Computing}, vol.~2, pp. 1--11. EasyChair (2012)

\bibitem{quaife1992}
Quaife, A.: Automated Development of Fundamental Mathematical Theories.
\newblock Springer, Berlin Heidelberg New York (1992)

\bibitem{schwabhauser}
Schwabh\"auser, W., Szmielew, W., Tarski, A.: {M}etamathematische {M}ethoden in
  der {G}eometrie: {T}eil {I}: {E}in axiomatischer {A}ufbau der euklidischen
  {G}eometrie, {T}eil {II}: {M}etamathematische {B}etrachtungen
  ({H}ochschultext).
\newblock Springer--Verlag (1983).
\newblock Reprinted 2012 by Ishi Press, with a new foreword by Michael Beeson.

\bibitem{tarski1951}
Tarski, A.: A decision method for elementary algebra and geometry.
\newblock Tech. Rep. R-109, second revised edition, reprinted in
  \cite{cavinessjohnson}, pp. 24--84, Rand Corporation (1951)

\bibitem{tarski1959}
Tarski, A.: What is elementary geometry?
\newblock In: L.~Henkin, P.~Suppes, A.~Tarksi (eds.) The axiomatic method, with
  special reference to geometry and physics. Proceedings of an International
  Symposium held at the {U}niv. of {C}alif., {B}erkeley, {D}ec. 26,
  1957--{J}an. 4, 1958, Studies in Logic and the Foundations of Mathematics,
  pp. 16--29. North-Holland, Amsterdam (1959).
\newblock Available as a 2007 reprint, Brouwer Press, ISBN 1-443-72812-8

\bibitem{tarski-givant}
Tarski, A., Givant, S.: Tarski's system of geometry.
\newblock The Bulletin of Symbolic Logic \textbf{5}(2), 175--214 (1999)

\bibitem{veblen1904}
Veblen, O.: A system of axioms for geometry.
\newblock Transactions of the American Mathematical Society \textbf{5},
  343--384 (1904)

\bibitem{veroff1996}
Veroff, R.: Using hints to increase the effectiveness of an automated reasoning
  program.
\newblock Journal of Automated Reasoning \textbf{16}(3), 223--239 (1996)

\bibitem{wos1988}
Wos, L.: Automated reasoning: 33 basic research problems.
\newblock Prentice Hall, Englewood Cliffs, New Jersey (1988)

\bibitem{wos2003}
Wos, L.: Automated reasoning and the discovery of missing and elegant proofs.
\newblock Rinton Press, Paramus, New Jersey (2003)

\bibitem{wos-notebook2008}
Wos, L.: The subformula strategy: coping with complex expressions\hfill (2008).
\newblock
  {h}ttp://www.automatedreasoning.net/docs\_and\_pdfs/subformula

\bibitem{wos-notebookTarski}
Wos, L.: An amazing approach to plane geometry\hfill (2014).
\newblock
  {h}ttp://www.automatedreasoning.net/docs\_and\_pdfs/an\_amazing\_approach

\bibitem{wos-fascinating}
Wos, L., Pieper, G.W.: A fascinating country in the world of computing.
\newblock World Scientific (1999)

\end{thebibliography}
 \end{document}